%% file: main.tex
\documentclass[runningheads,a4paper]{llncs}
\usepackage[top=2.5cm, bottom=2.5cm, left=2.5cm, right=2.5cm]{geometry}
\usepackage{amsmath}
\usepackage{amssymb}
\usepackage{graphicx}
\usepackage{booktabs}
\usepackage{listings}
\usepackage{xcolor}
\usepackage{hyperref}
\usepackage{caption}
\usepackage{subcaption}

\usepackage{url}
\usepackage{color}
\usepackage{cite}
\usepackage{epsfig}

\usepackage[nomargin,inline,marginclue,draft]{fixme}
\usepackage{cleveref}
\fxsetup{status=draft}
\fxsetup{theme=color, mode=multiuser} \FXRegisterAuthor{me}{ame}{\color{red}Me}
\newcommand{\orcid}[1]{\href{https://orcid.org/#1}{\includegraphics[scale=0.02]{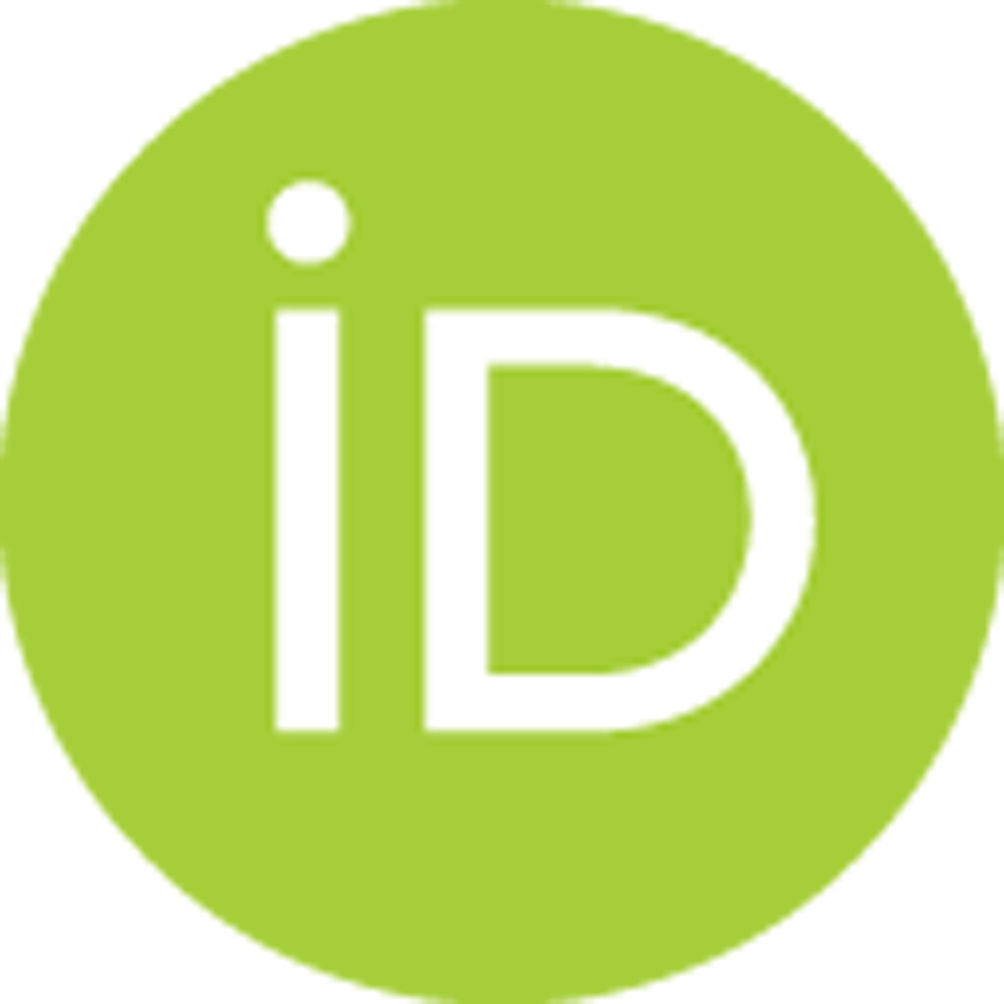}}} 

\hypersetup{
    colorlinks=true
}
\title{The Brain Tumor Segmentation in Pediatrics (BraTS-PEDs) Challenge: \emph{Focus on Pediatrics (CBTN-CONNECT-DIPGR-ASNR-MICCAI BraTS-PEDs)}}
\titlerunning{The BraTS-PEDs Challenge}

\input{0_author_list}

\begin{document}
    \mainmatter
    \maketitle
    \setcounter{footnote}{0} 
     \begin{abstract}
         Pediatric tumors of the central nervous system are the most common cause of cancer-related death in children. The five-year survival rate for high-grade gliomas in children is less than 20\%. Due to their rarity, the diagnosis of these entities is often delayed, their treatment is mainly based on historic treatment concepts, and clinical trials require multi-institutional collaborations. Here we present the CBTN-CONNECT-DIPGR-ASNR-MICCAI BraTS-PEDs challenge, focused on pediatric brain tumors with data acquired across multiple international consortia dedicated to pediatric neuro-oncology and clinical trials. 
         The CBTN-CONNECT-DIPGR-ASNR-MICCAI BraTS-PEDs challenge brings together clinicians and AI/imaging scientists to lead to faster development of automated segmentation techniques that could benefit clinical trials, and ultimately the care of children with brain tumors.
    \end{abstract}
    
    
    \input{1_introduction}
         
    \section{Challenge Design}
        \input{2_Material_and_Methods.tex}

        \input{4_discussion}

    \section*{Acknowledgments}
        Success of any challenge in the medical domain depends upon the quality of well annotated multi-institutional datasets. We are grateful to all the data contributors, annotators, and approvers for their time and efforts. Our profound thanks go to the Children's Brain Tumor Network (CBTN), the Collaborative Network for Neuro-oncology Clinical Trials (CONNECT), the International DIPG/DMG Registry (DIPGR), the American Society of Neuroradiology (ASNR), and the Medical Image Computing and Computer Assisted Intervention (MICCAI) Society for their invaluable support of this challenge.
    
    \section*{Funding}
    
    Research reported in this publication was partly supported by the National Institutes of Health (NIH) under award numbers: NCI/ITCR:U01CA242871 and NCI:UH3CA236536, and by grant funding from the Pediatric Brain Tumor Foundation and DIPG/DMG Research Funding Alliance (DDRFA). The content of this publication is solely the responsibility of the authors and does not represent the official views of the NIH.

    \bibliographystyle{ieeetr}
    \bibliography{bibliography.bib}
    \newpage
    \appendix
\end{document}

%% file: 0_author_list.tex
\author{
Anahita Fathi Kazerooni \inst{1,2,3,\dag,\ddag,\orcid{0000-0001-7131-2261}}
\and Nastaran Khalili \inst{1,\dag, \S,\orcid{0000-0001-8078-3591}}
\and Xinyang Liu \inst{4,\ddag}
\and Deep Gandhi \inst{1,\dag}
\and Zhifan Jiang \inst{4,\ddag}
\and Syed Muhammed Anwar \inst{4,\ddag}
\and Jake Albrecht \inst{5, \dag}
\and Maruf Adewole \inst{6,\dag}
\and Udunna Anazodo \inst{7,\dag}
\and Hannah Anderson \inst{8, \S}
\and Ujjwal Baid \inst{3,8,9,\dag,\orcid{0000-0001-5246-2088}}
\and Timothy Bergquist \inst{5,\dag}
\and Austin J. Borja \inst{10, \S}
\and Evan Calabrese \inst{11,\dag}
\and Verena Chung \inst{5,\dag}
\and Gian-Marco Conte \inst{12,\dag}
\and Farouk Dako \inst{13,\dag}
\and James Eddy \inst{5,\dag}
\and Ivan Ezhov \inst{14,15,\dag}
\and Ariana Familiar \inst{1}
\and Keyvan Farahani \inst{16,\dag}
\and Andrea Franson \inst{17, \dag}
\and Anurag Gottipati \inst{1,\dag}
\and Shuvanjan Haldar \inst{18, \S}
\and Juan Eugenio Iglesias \inst{19,\dag}
\and Anastasia Janas \inst{20,\dag}
\and Elaine Johansen \inst{21,\dag}
\and Blaise V Jones \inst{22, \S, \orcid{0000-0001-9003-4578}}
\and Neda Khalili \inst{1,\dag, \S}
\and Florian Kofler \inst{23,\dag}
\and Dominic LaBella \inst{24,\dag}
\and Hollie Anne Lai \inst{25, \S, \orcid{0000-0001-6747-7338}}
\and Koen Van Leemput \inst{26,\dag}
\and Hongwei Bran Li \inst{19,\dag}
\and Nazanin Maleki \inst{20,\ddag}
\and Aaron S McAllister \inst{27, \S, \orcid{0000-0002-1117-9355}}
\and Zeke Meier \inst{28,\dag}
\and Bjoern Menze \inst{29,30,\dag}
\and Ahmed W Moawad \inst{31,\dag}
\and Khanak K Nandolia \inst{32, \S, \orcid{0000-0002-8627-2992}}
\and Julija Pavaine \inst{33, 34, \S, \orcid{0000-0003-1843-837X}}
\and Marie Piraud \inst{23,\dag}
\and Tina Poussaint \inst{35, \ddag}
\and Sanjay P Prabhu \inst{36, \S, \orcid{0000-0003-0871-115X}}
\and Zachary Reitman \inst{24,\dag}
\and Jeffrey D Rudie \inst{37,\dag,\orcid{0000-0000-0000-0000}}
\and Mariana Sanchez-Montano \inst{38, \S, \orcid{0000-0002-9203-1060}}
\and Ibraheem Salman Shaikh \inst{39, \S, \orcid{0000-0002-5994-997}} 
\and Nakul Sheth \inst{40, \S} 
\and Wenxin Tu \inst{41,\S} 
\and Chunhao Wang \inst{24,\dag}
\and Jeffrey B Ware \inst{1,\S}
\and Benedikt Wiestler \inst{30,\dag}
\and Anna Zapaishchykova \inst{35, \ddag}
\and Miriam Bornhorst \inst{42,\ddag,\orcid{0000-0000-0000-0000}} 
\and Michelle Deutsch \inst{27,\ddag}
\and Maryam Fouladi \inst{27,\ddag}
\and Margot Lazow \inst{27,\ddag}
\and Leonie Mikael \inst{27,\ddag}
\and Trent Hummel \inst{43, \ddag} 
\and Benjamin Kann \inst{35, \ddag} 
\and Peter de Blank \inst{43, \ddag} 
\and Lindsey Hoffman \inst{44}
\and Mariam Aboian \inst{40,\dag, \ddag,\orcid{0000-0002-4877-8271}}
\and Ali Nabavizadeh \inst{1,8,\ddag,\S,\orcid{0000-0002-0380-4552}}
\and Roger Packer \inst{42,\ddag}
\and Spyridon Bakas \inst{3,8,9,\dag,\orcid{0000-0001-8734-6482}}
\and Adam Resnick \inst{1,2,\ddag,\orcid{0000-0003-0436-4189}}
\and Brian Rood \inst{45,\ddag}
\and Arastoo Vossough \inst{1,40,\dag,\S, \orcid{0000-0003-1346-427X}}
\and Marius George Linguraru \inst{4,46,\dag,\ddag,**, \orcid{0000-0001-6175-8665}}
}

\authorrunning{Kazerooni A, et al.}

\institute{\scriptsize{ Center for Data-Driven Discovery in Biomedicine (D3b), Children's Hospital of Philadelphia, Philadelphia, PA, USA
\and Department of Neurosurgery, University of Pennsylvania, Philadelphia, PA, USA
\and Center for AI \& Data Science for Integrated Diagnostics (AI2D) and Center for Biomedical Image Computing and Analytics (CBICA), University of Pennsylvania, Philadelphia, PA, USA
\and Sheikh Zayed Institute for Pediatric Surgical Innovation, Children's National Hospital, Washington DC, USA
\and Sage Bionetworks, USA
\and Medical Artificial Intelligence (MAI) Lab, Crestview Radiology, Lagos, Nigeria
\and Montreal Neurological Institute (MNI), McGill University, Montreal, QC, Canada
\and Department of Radiology, Perelman School of Medicine, University of Pennsylvania, Philadelphia, PA, USA
\and Division of Computational Pathology, Department of Pathology and Laboratory Medicine, Indiana University School of Medicine, Indianapolis, IN, USA
\and Department of Neurosurgery at the University of Southern California, CA, USA
\and Department of Radiology, Duke University Medical Center, Durham, NC, USA
\and Mayo Clinic, MN, USA
\and Center for Global Health, Perelman School of Medicine, University of Pennsylvania, Philadelphia, PA, USA
\and Department of Informatics, Technical University Munich, Germany
\and TranslaTUM - Central Institute for Translational Cancer Research, Technical University of Munich, Germany
\and Cancer Imaging Program, National Cancer Institute, National Institutes of Health, Bethesda, MD, USA
\and C. S. Mott Children's Hospital, University of Michigan, MI, USA
\and Biomedical Engineering Rutgers University, New Brunswick, NJ, USA
\and Athinoula A Martinos Center for Biomedical Imaging, Massachusetts General Hospital, Boston, MA, USA
\and Yale University, New Haven, CT, USA
\and PrecisionFDA, U.S. Food and Drug Administration, Silver Spring, MD, USA
\and Cincinnati Children’s Hospital Medical Center, OH, USA (ORCID ID: 0000-0001-9003-4578)
\and Helmholtz AI, Helmholtz Munich, Germany
\and Department of Radiation Oncology, Duke University Medical Center, Durham, NC, USA
\and Department of Radiology, Children’s Health Orange County, CA, USA (ORCID: 0000-0001-6747-7338)
\and Department of Applied Mathematics and Computer Science, Technical University of Denmark, Denmark
\and Department of Radiology, Nationwide Children's Hospital, OH, USA 
\and Booz Allen Hamilton, McLean, VA, USA
\and Biomedical Image Analysis \& Machine Learning, Department of Quantitative Biomedicine, University of Zurich, Switzerland
\and Department of Neuroradiology, Technical University of Munich, Munich, Germany
\and Mercy Catholic Medical Center, Darby, PA, USA
\and Department of Diagnostic and Interventional Radiology, All India Institute of Medical Sciences, Rishikesh, India (ORCID: 0000-0002-8627-2992)
\and Department of Paediatric Radiology, Royal Manchester Children’s Hospital, Manchester University Hospitals NHS Foundation Trust, University of Manchester, Manchester, UK (ORCID: 0000-0003-1843-837X)
\and Division of Informatics, Imaging \& Data Sciences, School of Health Sciences, Faculty of Biology, Medicine and Health, University of Manchester, Manchester Academic Health Science Centre, Manchester, UK
\and Dana-Farber Brigham Cancer Center \& Boston Children’s Hospital, Boston, MA, USA
\and Division of Neuroradiology, Department of Radiology, Boston Children's Hospital, Harvard Medical School, Boston, Massachusetts, USA  (ORCID: 0000-0003-0871-115X)
\and University of San Diego, CA, USA
\and UNIDAD DE PATOLOGIA CLINICA:Guadalajara, JALISCO, Mexico (ORCID: 0000-0002-9203-1060) 
\and Beth Israel Deaconess Medical Center, Harvard Medical School, MA, USA (ORCID: 0000-0002-5994-997)
\and Department of Radiology, The Children's Hospital of Philadelphia, PA, USA
\and College of Arts and Sciences, University of Pennsylvania, PA, USA
\and Brain Tumor Institute, Children's National Hospital, Washington, DC, USA
\and Brain Tumor Center, Cincinnati Children's Hospital, Cincinnati, OH, USA
\and Center for Cancer and Blood Disorders, Phoenix Children's Hospital, Phoenix, AZ, US
\and Center for Cancer and Blood Disorders, Children’s National Hospital, Washington, DC, USA
\and Departments of Radiology and Pediatrics, George Washington University School of Medicine and Health Sciences, Washington, DC, USA
}
\linebreak
\\
\textsuperscript{\dag} People involved in the organization of the challenge.\\
\textsuperscript{\ddag} People contributing data from their institutions.\\
\textsuperscript{\S} People involved in annotation process.\\ 
\textsuperscript{**} Corresponding author: \email{\{mlingura@childrensnational.org\}}}

%% file: 1_introduction.tex
\section{Introduction}

   Pediatric diffuse midline gliomas (DMGs, including pediatric diffuse intrinsic pontine glioma (DIPGs)) are high grade gliomas with short average overall survival \cite{mackay2017integrated, jansen2015survival}. Many DMGs are located in the pons and often diagnosed between 5 and 10 years of age. Pediatric brain tumors require dedicated tumor segmentation tools that help in their characterization and facilitate their diagnosis, prognosis, and treatment response assessment \cite{fathi2023automated,madhogarhia2022radiomics,nabavizadeh2023current}. There are only a handful of automated tumor segmentation methods explicitly proposed for pediatric brain tumors \cite{fathi2023automated, nalepa2022segmenting, artzi2020automatic, tor2020unsupervised, mansoor2016deep, avery2016optic, mansoor2017joint, peng2022deep, vossough2024training, liu2023, Boyd2023.06.29.23292048}. However, the majority of these methods have been developed only for the segmentation of the T2 fluid attenuated inversion recovery (FLAIR) abnormal signal \cite{nalepa2022segmenting,artzi2020automatic,peng2022deep}, also called whole tumor (WT). 
      
   The MICCAI brain tumor segmentation (BraTS) challenges have established a community benchmark dataset and environment for adult glioma over the past 12 years  \cite{menze2014multimodal,bakas2017advancing,bakas2018identifying,baid2021rsna}. In 2023 challenge, we launched the first Brain Tumor Segmentation in Pediatrics (BraTS-PEDs) challenge. In 2024, we continue the BraTS-PEDs challenge with modifications in processing pipeline and tumor subregions. We have created a retrospective multi-institutional (multi-consortium) pediatric database, with the data collected through a few consortia, including Children’s Brain Tumor Network (CBTN, \url{https://cbtn.org/} ) \cite{lilly2023children}, DIPG Registry (\url{https://www.dipgregistry.org}) and the COllaborative Network for NEuro-oncology Clinical Trials (CONNECT, \url{https://connectconsortium.org/}). Additional data from participating pediatric institutions have been included in the BraTS-PEDs cohort. The American Society of Neuroradiology (ASNR, \url{https://www.asnr.org/}) collaborated in generating ground truth annotation for the majority of data in this challenge.  This manuscript provides an overview of the CBTN-CONNECT-ASNR-MICCAI BraTS-PEDs challenge.

%% file: 2_Material_and_Methods.tex
\subsection{Data}

    The BraTS-PEDs dataset includes a retrospective multi-institutional cohort of conventional/structural magnetic resonance imaging (MRI) sequences, including pre- and post-gadolinium T1-weighted (labeled as T1 and T1CE), T2-weighted (T2), and T2-weighted fluid attenuated inversion recovery (T2-FLAIR) images, from 464 pediatric high-grade glioma. These conventional multiparametric MRI (mpMRI) sequences are commonly acquired as part of standard clinical imaging for brain tumors. However, the image acquisition protocols and MRI equipment differ across different institutions, resulting in heterogeneity in image quality in the provided cohort. Inclusion criteria comprised of pediatric subjects with: (1) histologically-approved high-grade glioma, i.e., high-grade astrocytoma and diffuse midline glioma (DMG), including radiologically or histologically-proven diffuse intrinsic pontine glioma (DIPG); (2) availability of all four structural mpMRI sequences on treatment-naive imaging sessions. Exclusion criteria consisted of: (1) images assessed to be of low quality or with artifacts that would not allow for reliable tumor segmentation; and (2) infants younger than one month of age. Data for 464 patients was obtained through CBTN (n = 120), DMG/ DIPG Registry (n = 256), Boston's Children Hospital (n = 61), and Yale University (n = 27). 
     
    The cohort included in the challenge is split into training, validation, and testing datasets. The data shared with the participants comprise mpMRI scans and ground truth labels for the training cohort, as well as mpMRI sequences without any associated ground truth for the validation cohort. Notably, the testing data that will be used for evaluating the performance of the methods submitted by challenge participants will not be shared with the participants, but the containerized submissions of the participants will be evaluated by the synapse.org platform, powered by MedPerf\cite{karargyris2021medperf}. 
     
    \textit{Participants are prohibited from training their algorithm on any additional public and/or private data (from their own institutions) besides the provided BraTS-PEDs data, or using models pretrained on any other dataset. This restriction is imposed to allow for a fair comparison among the participating methods. However, participants can use additional public and/or private data only for publication of their scientific papers, if they also provide a report of their results on the data from the BraTS-PEDs challenge alone, and discuss potential differences in the obtained results.}

\label{sec:data}

    \subsubsection{Imaging Data Description}  

    For all patients, mpMRI scans were prepared using the following steps:
    \begin{enumerate}
    \item Pre-processing using the "BraTS Pipeline", a standardized approach, publicly available through the Cancer Imaging Phenomics Toolkit (CaPTk) \footnote{\url{https://cbica.github.io/CaPTk/}} \cite{captk,captk_2,captk_3} and Federated Tumor Segmentation (FeTS) tool \footnote{\url{https://github.com/FETS-AI/Front-End/}}. De-identification was performed through removing protected Health Information (PHI) from DICOM headers in DICOM to NIfTI conversion step \cite{NEJMc1908881,NEJMc1915674}. 
    \item Defacing using a pediatric-specific automated defacing method \footnote{\url{https://github.com/d3b-center/peds-auto-defacing-public}}
    \item Tumor subregion segmentation using a pediatric autosegmentation method \cite{vossough2024training} \footnote{\url{https://github.com/d3b-center/peds-brain-seg-pipeline-public/}}
    \end{enumerate}
         
   The result of this step was segmentation of the tumors into four main subregions, recommended by RAPNO working group for evaluation of the treatment response in high-grade gliomas and DIPGs. The annotated tumor subregions comprised of the following regions \cite{fathi2023automated}:
    \begin{enumerate}
    \item Enhancing tumor (ET; label 1; value = 1), described by areas with enhancement (brightness) on T1 post-contrast images as compared to T1 pre-contrast. In case of mild enhancement, checking the signal intensity of normal brain structure can be helpful.
   \item  Nonenhancing tumor (NET; label 2; value = 2), defined as any other abnormal signal intensity within the tumorous region that cannot be defined as enhancing or cystic. For example, the abnormal signal intensity on T1, T2-FLAIR, and T2 that is not enhancing on T1CE should be considered as nonenhancing portion. 
   \item  Cystic component (CC; label 3; value = 3), typically appearing with hyperintense signal (very bright) on T2 and hypointense signal (dark) on T1CE. The cystic portion should be within the tumor, either centrally or peripherally (as compared to ED which is peritumoral). The brightness of CC is here defined as comparable or close to cerebrospinal fluid (CSF).
   \item  Peritumoral edema (ED; label 4; value = 4), defined by the abnormal hyperintense signal (very bright) on FLAIR scans. ED is finger-like spreading that preserves underlying brain structure and surrounds the tumor. 

   The BraTS-PEDs 2024 can be accessed at \url{https://www.synapse.org/Synapse:syn58894466}.
   \end{enumerate}

   \textbf{Important Note: The BraTS-PEDs 2024 is not skull-stripped}. If you need to skull strip the images, you may use our open-access model provided at one of the following repositories:
     \begin{enumerate} 
     \item \url{https://github.com/d3b-center/peds-brain-seg-pipeline-public/}
     \item \url{https://github.com/d3b-center/peds-brain-auto-skull-strip}

      We caution the participants about using skull-stripping in training their models, as the validation and testing data are only defaced and not skull-stripped. 
     \end{enumerate}

   The automatically-generated four labels (Figure 1) using the automated segmentation tool were used as preliminary segmentation to be manually revised by volunteer neuroradiology experts of varying rank and experience, in accordance with annotation guidelines.  
    The volunteer neuroradiology expert annotators were provided with four mpMRI sequences (T1, T1CE, T2, FLAIR) along with the fused automated segmentation volume to initiate the manual refinements. The ITK-SNAP \cite{itksnap} software was used for making these refinements. Once the automated segmentations were refined by the annotators, three  attending board-certified neuroradiologists, reviewed the segmentations. Depending upon correctness, these segmentations were either approved or returned to the individual annotator for further refinements. This process was followed iteratively until the approvers found the refined tumor subregion segmentations acceptable for public release and the challenge conduction.

    \begin{figure}[t]
          \centering
          \includegraphics[width=0.9\linewidth]{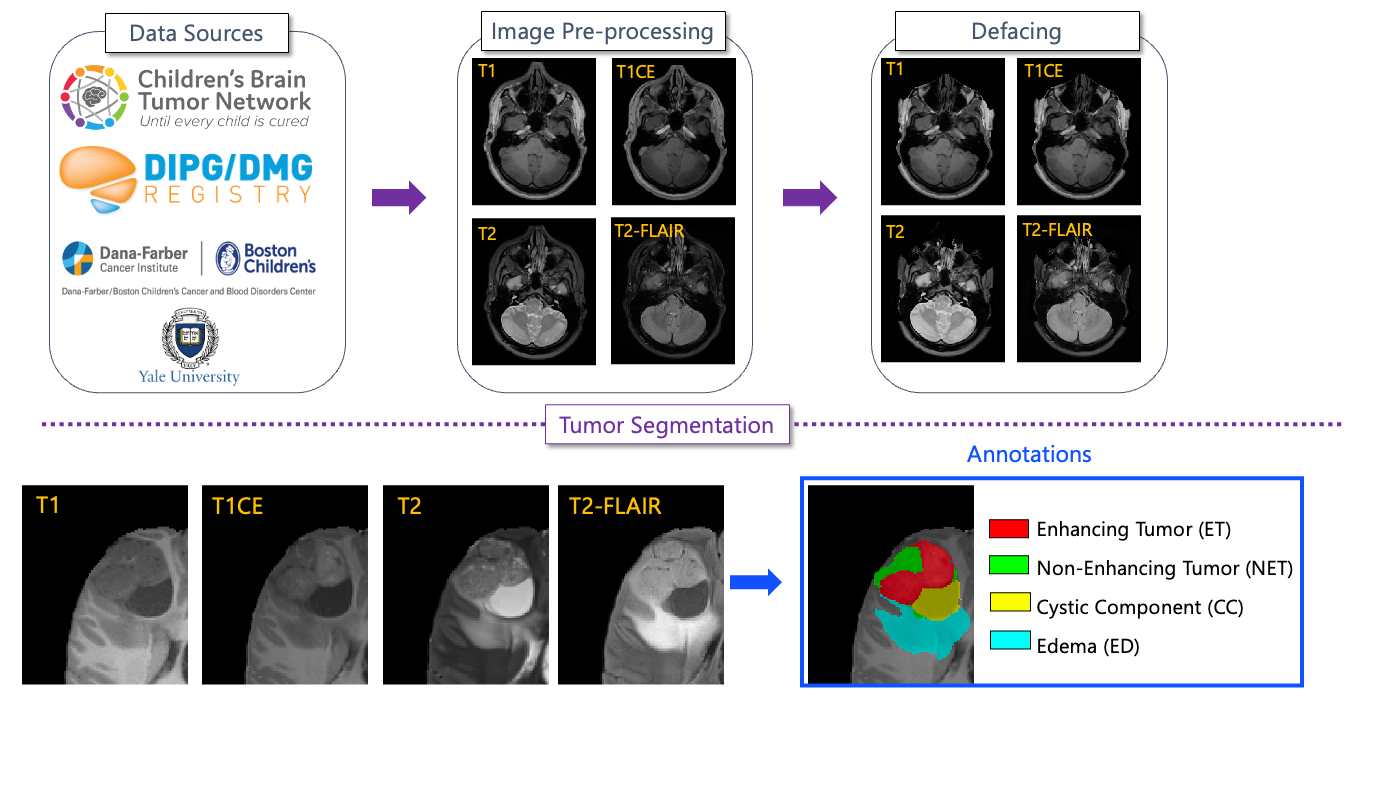}  
          \caption{\textbf{Graphical representation of data processing and annotations in pediatric brain tumors.} Top panel presents the processing pipeline, and the bottom panel illustrates the annotated tumor subregions along with mpMRI structural scans (T1, T1CE, T2, and T2-FLAIR). Tumor subregions include the enhancing tumor (ET - red), non-enhancing tumor (NET - green), cystic component (CC - yellow), and edema (ED - teal) regions. }
        \label{annotations}
    \end{figure}

    \subsubsection{Common errors of automated segmentations}

     In our experience with automated segmentation of pediatric brain tumors, some errors may be noticed:
            
            \begin{enumerate}
                \item  Peri-ventricular edema  segmented as ED
                \item  Remote areas segmented as tumor (far from actual tumor region)
                \item  Under-segmentation of cysts
                \item  Non-enhancing tumor segmented as cyst, or vice versa: if there is an enhancing rim around a cyst-looking portion, this is considered NET. If cyst-looking portion is very bright on T2 and dark on T1, then it is cyst.
            \end{enumerate}

    \subsection{Performance Evaluation}
\textbf{Important Note: The BraTS-PEDs 2024 evaluation is different from the 2023 challenge.}

    For the BraTS-PEDs 2024 challenge, the regions to evaluate the performances are the: i) ``enhancing tumor'' (ET), ii) ``non-enhancing tumor'' (NET), iii) ``cystic component'' (CC), iv) ``edema'' (ED), v) ``tumor core'' (TC) as a combination of ET, NET, and CC, vi) the entire tumorous region, or the so-called ``whole tumor'' (WT).

    The participants are required to send the output of their methods to the evaluation platform for the scoring to occur during the training and the validation phases. At the end of the validation phase the participants are asked to identify the method they would like to evaluate in the final testing/ranking phase. The organizers will then confirm receiving the containerized method and will evaluate it on withheld testing data. The participants will be provided guidelines on the form of the container as we have done in previous years. This will enable confirmation of reproducibility, comparing these algorithms to the previous BraTS instances and comparison with results obtained by algorithms of previous years, thereby maximizing solutions in solving the problem of brain tumor segmentation. 
    
    During the training and validation phases, the participants will be able to test the functionality of their submission through three platforms:
            \begin{enumerate}
                \item Cancer Imaging Phenomics Toolkit (CaPTk [5-6], \url{https://github.com/CBICA/CaPTk} )
                \item Federated Tumor Segmentation (FeTS) Tool [7] (\url{https://fets-ai.github.io/Front-End/})
                \item Online evaluation platform (Synapse \url{https://www.synapse.org/Synapse:syn53708249/wiki/626323})
            
             \end{enumerate} 
    
    \subsection{Participation Timeline}  
        
        The challenge is composed of three main stages:
                    
            \begin{enumerate}
                \item  Training: The four MRI sequences along with the corresponding ground truth labels will be shared with participants to design and train their methods.
                \item  Validation: The validation data will be released to the participants within three weeks after the training data. The participants will not be provided with the ground truth of the validation data, but will be given the opportunity to submit multiple times to the online evaluation platforms. The participants can generate preliminary results for their trained models in unseen data and report them in their submitted short MICCAI LNCS papers , in addition to their cross-validated results on the training data, to be published in conjunction with the proceedings of the BrainLes workshop. The top-ranked participating teams in the validation phase will be invited to prepare their slides for a short oral presentation of their method during the BraTS challenge at MICCAI 2024. 
                \item  Testing/Ranking: After the participants upload their containerized method in the evaluation platforms, they will be evaluated and ranked on unseen testing data, which will not be made available. The final top-ranked participating teams will be announced at the 2024 MICCAI Annual Meeting. 
            \end{enumerate}

%% file: 4_discussion.tex
\section{Discussion}

In this paper, we outlined the design of BraTS-PEDs challenge, to benchmark methods devised for segmentation of pediatric brain tumors.
We are actively working on increasing the number of subjects in this cohort to provide the community with a large dataset of these rare tumors, and to facilitate the future development of tools for computer-aided treatment planning. 
Future BraTS-PEDs challenges will include data from more institutions and tumor histologies, and will be extended to post-operative or post-treatment scans.